\documentclass[sigconf]{acmart}

\renewcommand\footnotetextcopyrightpermission[1]{} 
\usepackage{balance}

\usepackage{graphicx}
\usepackage{url}
\usepackage{ amssymb }
\usepackage{color,soul}

\usepackage[T1]{fontenc}
\usepackage[scaled=0.85]{beramono}
\usepackage{listings}

\usepackage{array}
\usepackage{booktabs}
\usepackage{multirow}

\begingroup\expandafter\expandafter\expandafter\endgroup
\expandafter\ifx\csname IncludeInRelease\endcsname\relax
\usepackage{fixltx2e}
\fi

\usepackage{graphicx}
\usepackage{pict2e}
\usepackage{subcaption}
\usepackage{amsmath}

\usepackage[colorinlistoftodos]{todonotes} 

\newcommand{\db}{\textit{DB18}}
\newcommand{\wiki}{\textit{WikiGeo19}}
\newcommand{\kg}{\mathcal{G}}
\newcommand{\entset}{\mathcal{V}}
\newcommand{\relset}{\mathcal{R}}
\newcommand{\triset}{\mathcal{T}}

\newcommand{\gp}{b}
\newcommand{\ent}{e}
\newcommand{\rel}{r}
\newcommand{\variable}{V}
\newcommand{\typefun}{\Gamma}
\newcommand{\type}{\gamma}
\newcommand{\embmat}{\mathbf{Z}}
\newcommand{\entemb}{\mathbf{e}}
\newcommand{\embdim}{d}

\newcommand{\numentpertype}{m_{\type}}
\newcommand{\realnum}{\mathbb{R}}
\newcommand{\onehotvec}{\mathbf{x}}
\newcommand{\cgq}{q}

\newcommand{\proj}{\mathcal{P}}
\newcommand{\projmat}{\mathbf{R}}

\newcommand{\inter}{\mathcal{I}}
\newcommand{\intermat}{\mathbf{W}}
\newcommand{\interbia}{\mathbf{B}}
\newcommand{\interelemfun}{\mathbf{\Psi}}

\newcommand{\attnfun}{\mathbf{A}}
\newcommand{\attnscr}{\alpha}
\newcommand{\attnvec}{\mathbf{a}}
\newcommand{\neisize}{n}
\newcommand{\attnsize}{K}
\newcommand{\attnidx}{k}
\newcommand{\attnact}{\sigma}
\newcommand{\lynorm}{LayerNorm}

\newcommand{\kgtembed}{\mathbf{H}_{KG}}

\newcommand{\kgtloss}{\mathcal{L}_{KG}}
\newcommand{\qatloss}{\mathcal{L}_{QA}}
\newcommand{\loss}{\mathcal{L}}

\newcommand{\cosine}{\mathbf{\Phi}}

\newcommand{\nei}{N}
\newcommand{\negsamp}{Neg}

\newcommand{\qasamplesize}{Q}

\newcommand{\qaset}{S}

\newcommand{\query}{q}
\newcommand{\answer}{a}

\newcommand{\queryembed}{\mathbf{q}}
\newcommand{\answerembed}{\mathbf{a}}

\newcommand{\relu}{ReLU}
\newcommand{\leakyrelu}{LeakyReLU}

\newcommand{\model}{CGA}
\newcommand{\baseline}{GQE}

\newcommand{\cga}[1]{\textbf{\model+KG#1[min]}}

\newcommand{\datathreshold}{$\eta$}

\def\BibTeX{{\rm B\kern-.05em{\sc i\kern-.025em b}\kern-.08emT\kern-.1667em\lower.7ex\hbox{E}\kern-.125emX}}
    
\copyrightyear{2019} 
\acmYear{2019} 
\setcopyright{acmcopyright}
\acmConference[K-CAP '19]{Proceedings of the 10th International Conference on Knowledge Capture}{November 19--21, 2019}{Marina Del Rey, CA, USA}
\acmBooktitle{Proceedings of the 10th International Conference on Knowledge Capture (K-CAP '19), November 19--21, 2019, Marina Del Rey, CA, USA}
\acmPrice{15.00}
\acmDOI{10.1145/3360901.3364432}
\acmISBN{978-1-4503-7008-0/19/11}

\begin{document}

\fancyhead{}

\title{Contextual Graph Attention for Answering Logical Queries over Incomplete Knowledge Graphs}

%
%

\author{Gengchen Mai, Krzysztof Janowicz, Bo Yan, Rui Zhu, Ling Cai, Ni Lao}

\affiliation{%
	\institution{STKO Lab, UCSB}
}
\email{{gengchen_mai, jano, boyan, ruizhu, lingcai}@geog.ucsb.edu}

\affiliation{%
	\institution{SayMosaic Inc.}
}
\email{ni.lao@mosaix.ai}

%
\renewcommand{\shortauthors}{Gengchen Mai et al.}

%
\begin{abstract}
	Recently, several studies have explored methods for using KG embedding to answer logical 
	queries. 
	These approaches either treat embedding learning and query answering as two separated learning tasks, 
	or fail to deal with the variability of contributions from different query paths. 
	We proposed to leverage a graph attention mechanism \cite{velivckovic2017graph} to handle the unequal contribution of different query paths.
	However, commonly used graph attention assumes that the center node embedding is provided, which is unavailable in this task since the center node is to be predicted.
	%
	To solve this problem  we propose a multi-head attention-based end-to-end logical query answering model, called Contextual Graph Attention model (CGA), which uses an initial neighborhood aggregation layer to generate the center embedding, and the whole model is trained jointly on the original KG structure as well as the sampled query-answer pairs.
	We also introduce two new datasets, \db\ and \wiki, which are rather large in size compared to the existing datasets and contain many more relation types, and use them to evaluate the performance of the proposed model. 
	Our result shows that the proposed CGA with fewer learnable parameters consistently outperforms the baseline models on both datasets as well as Bio~\cite{hamilton2018embedding} dataset. 
\end{abstract}

%
%

%

%

%
\maketitle
\thispagestyle{empty}

\section{Introduction} \label{sec:intro}

Knowledge graphs 
represent statements in the form of graphs
in which nodes represent entities and directed labeled edges indicate different types of relations between these entities \cite{mai2018support}. In the past decade, the Semantic Web community has published and interlinked vast amounts of data on the Web using the machine-readable and reasonable Resource Description Framework (RDF) in order to create smart data \cite{janowicz2015data}. By following open W3C standards or related proprietary technology stacks, several large-scale knowledge graphs have been constructed (e.g., DBpedia, Wikidata, NELL, Google's Knowledge Graph and Microsoft's Satori) to support applications such as information retrieval and question answering \cite{berant2013semantic,liang2017neural}. 

Despite their size, knowledge graphs often suffer from incompleteness, sparsity, and noise as most KGs are constructed collaboratively and semi-automatically \cite{xu2016knowledge}. 
Recent work studied different ways of applying graph learning methods to large-scale knowledge graphs to support completion via so-called knowledge graph embedding techniques such as 
RESCAL \cite{nickel2012factorizing}, 
TransE \cite{bordes2013translating}, 
NTN \cite{socher2013reasoning}, 
DistMult \cite{yang2014embedding}, 
TransR \cite{lin2015learning}, and 
HOLE \cite{nickel2016holographic}. 
These approaches aim at embedding KG components including entities and relations into continuous vector spaces while preserving the inherent structure of the original KG \cite{wang2017knowledge}. Although these models show promising results in link prediction and entity classification tasks, they all treat each statement (often called \textit{triple}) independently, thereby ignoring the correlation between them. In addition, since the model needs to rank all entities for a given triple in the link prediction task, their complexity is linear with respect to the total number of entities in the KG, which makes it impractical for more complicated query answering tasks.

\begin{figure}[]
	\centering
	\setlength{\unitlength}{0.1\columnwidth}
	\begin{picture}(10,8)
    	\put(1,5.2){\includegraphics[width=0.8\columnwidth]{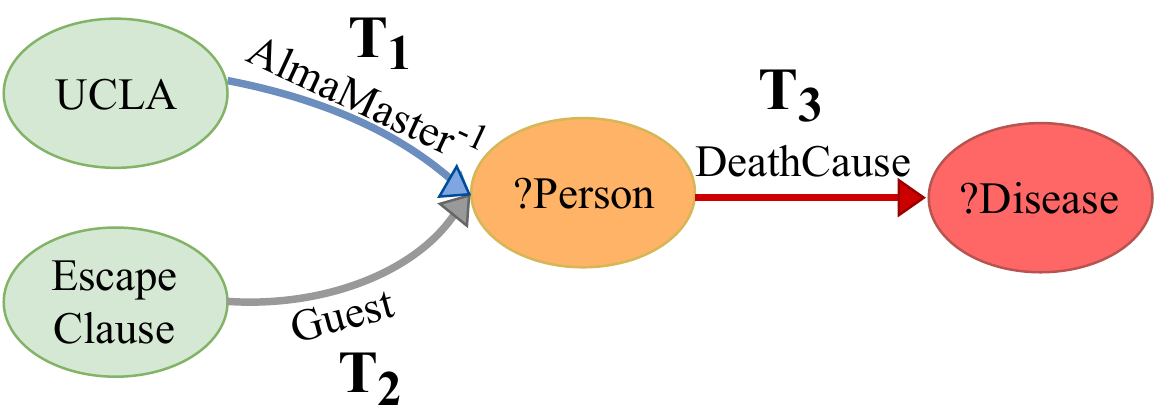}}

    	\put(1.5,4.8){$?Disease . \exists ?Person: AlmaMater^{-1}(UCLA, ?Person) \land $}
    	\put(4.7,4.4){$Guest(Escape Clause, ?Person) \land $}
    	\put(4.7,4.0){$DeathCause(?Person, ?Disease) $}
    	\put(0,3.9){\framebox(10,4.5){}}
    	 \put(1,0){\includegraphics[width=0.8\columnwidth]{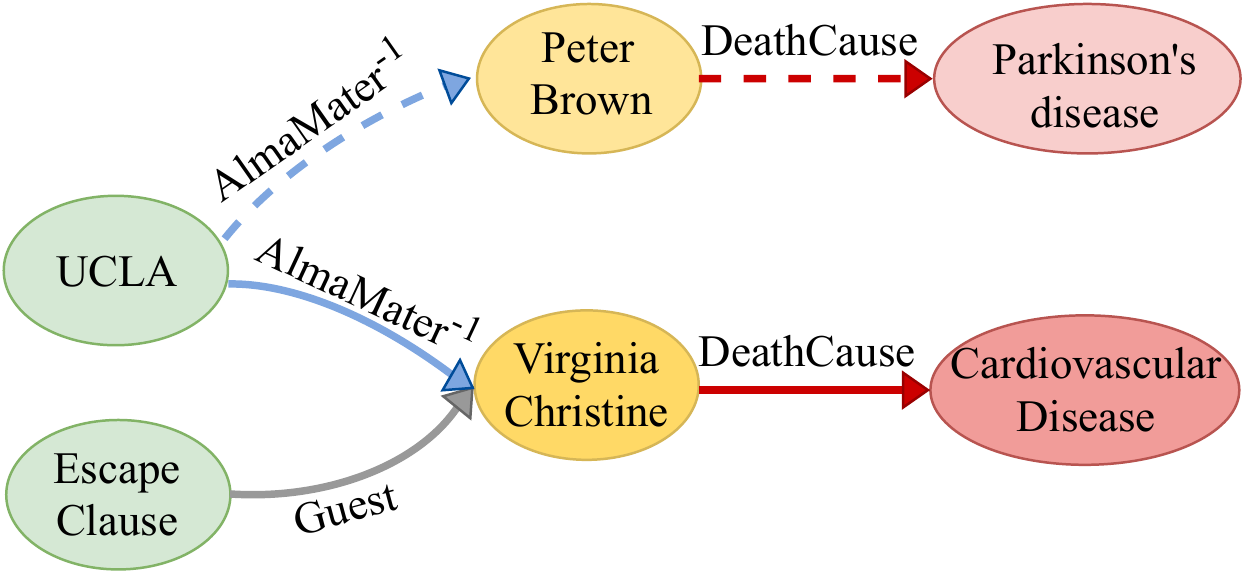}}
	\end{picture}
	\caption{
	\textbf{Top box:}  
	Conjunctive Graph Query (CGQ) and DAG of the query structure.
	\textbf{Below:}
	the matched underlining KG patterns represented by solid arrows.
	}
	\label{fig:qg}
	\vspace{-0.45cm}
\end{figure}
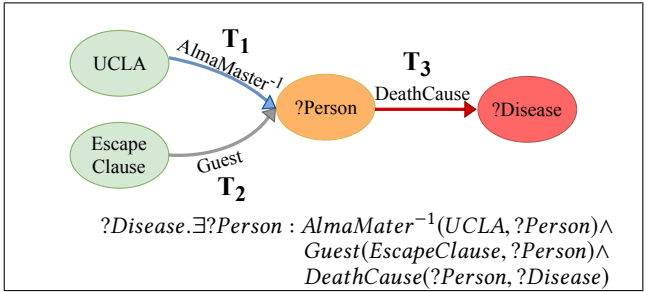

Recent work \cite{hamilton2018embedding,wang2018towards,mai2019relaxing} has explored ways to utilize knowledge graph embedding models for 
answering logical queries from incomplete KG.
The task is to predict the correct answer to a query based on KG embedding models, even if this query cannot be answered directly because of one or multiple missing triples in the original graph. For example, Listing \ref{q:exp} shows an example SPARQL query over DBpedia which asks for the cause of death of a person whose alma mater was UCLA and who was a guest of Escape Clause. Executing this query via DBpedia SPARQL endpoint\footnote{\url{https://dbpedia.org/sparql}} yields one answer \path{dbr:Cardiovascular_disease} and the corresponding person is \path{dbr:Virginia_Christine}. However, if the triple (\path{dbr:Virginia_Christine} \path{dbo:deathCause} \path{dbr:Cardiovascular_disease}) is missing, this query would become an unanswerable one \cite{mai2019relaxing}
as shown in Figure \ref{fig:qg}. 
%
The general idea of query answering via KG embedding is to predict the embedding of the root variable \textit{?Disease} by utilizing the embeddings of known entities (e.g. \path{UCLA} and \path{Escape Clause}) and relations (\path{deathCause}, \path{almaMater} and \path{guest}) in the query. Ideally, a nearest neighbor search in the entity embedding space using the predicted variable's embedding yields the approximated answer.

\vspace*{0.3cm}
\hspace*{-\parindent}%
\begin{minipage}[c]{\columnwidth}
	\begin{lstlisting}[
	linewidth=\columnwidth,
	breaklines=true,
	basicstyle=\ttfamily, 
	captionpos=b, 
	caption={An example SPARQL query over DBpedia}, 
	label={q:exp},
	frame=single
	]
	SELECT ?Disease 
	WHERE {
	?Person dbo:deathCause ?Disease. 
	?Person dbo:almaMater dbr:University_of_California,_Los_Angeles .  
	dbr:Escape_Clause dbo:guest ?Person .
	}
	\end{lstlisting}
\end{minipage}

Hamilton et al.~\cite{hamilton2018embedding} and Wang et al.~\cite{wang2018towards} proposed different approaches for predicting  variable embedding. However, an unavoidable step for both is to \textit{integrate} predicted embeddings for the same variable (in this query \textit{?Person}) from different paths (triple $T_{1}$ and $T_{2}$ in Fig. \ref{fig:qg}) by translating from the corresponding entity nodes via different relation embeddings. In Figure \ref{fig:qg}, triple $T_{1}$ and $T_{2}$ will produce different embeddings $\mathbf{p_{1}}$ and $\mathbf{p_{2}}$ for variable \textit{?Person} and they need to be integrated to produce one single embedding $\mathbf{p}$ for \textit{?Person}. An intuitive integration method is an element-wise mean operation over $\mathbf{p_{1}}$ and $\mathbf{p_{2}}$. This implies that we assume triple $T_{1}$ and $T_{2}$ have equal prediction abilities for the embedding of \textit{?Person} which is not necessarily true. In fact, triple $T_{1}$ matches 450 triples in DBpedia while $T_{2}$ only matches 5. This indicates that $\mathbf{p_{2}}$ will be more similar to the real embedding of \textit{?Person} because $T_{2}$ has more discriminative power. 

Wang et al.~\cite{wang2018towards} acknowledged this unequal contribution from different paths and obtained the final embedding $\mathbf{p}$ as a weighted average of $\mathbf{p_{1}}$ and $\mathbf{p_{2}}$ while the weight is proportional to the inverse of the number of triples matched by triple $T_{1}$ and $T_{2}$. However, this deterministic weighting approach lacks flexibility and will produce suboptimal results. Moreover, they separated the knowledge graph embedding training and query answering steps. As a result, the KG embedding model is not directly optimized on the query answering objective which further impacts the model's  performance. 

In contrast, Hamilton et al.~\cite{hamilton2018embedding} presented an end-to-end model for KG embedding model training and logical query answering. However, they utilized a simple permutation invariant neural network \cite{zaheer2017deep} to \textit{integrate} $\mathbf{p_{1}}$ and $\mathbf{p_{2}}$ which treats each embedding equally. Furthermore, in order to train the end-to-end logical query answering model, they sampled logical query-answer pairs from the KG as training datasets while ignoring the original KG structure which has proven to be important for embedding model training based on previous research \cite{kipf2016semi}.

Based on these observations, we hypothesis that a graph attention network similar to the one proposed by Veli{\v{c}}kovi{\'c} et al. \cite{velivckovic2017graph} can handle these unequal contribution cases. 
However,  Veli{\v{c}}kovi{\'c} et al. \cite{velivckovic2017graph}  assume that the center node embedding (the variable embedding of \textit{?Person} in Fig. \ref{fig:qg}), known as the \textit{query embedding} \cite{vaswani2017attention}, should be known beforehand for attention score computing which is unknown in this case. This prevents us from using the normal attention method.
Therefore, we propose an end-to-end attention-based logical query answering model over knowledge graphs in which the situation of unequal contribution from different paths to an entity embedding is  handled by a new attention mechanism \cite{bahdanau2014neural,vaswani2017attention,velivckovic2017graph} where \textbf{the center variable embedding is no longer a prerequisite}. Additionally, the model is jointly trained on both sampled logical query-answer pairs and the original KG structure information. 
\textbf{The contributions of our work are as follows:}
\begin{enumerate}
	\item We propose an end-to-end attention-based logic query answering model over knowledge graphs in which an attention mechanism is used to handle the unequal contribution of neighboring entity embeddings to the center entity embedding. 
	To the best of our knowledge, 
	this is the first attention method applicable to logic query answering.
	\item We show that the proposed model can be trained jointly on the original KG structure and the sampled logical QA pairs.
	\item We introduce two datasets - \db\ and \wiki\ -
	which have substantially more relation types (170+) compared to the Bio dataset \cite{hamilton2018embedding}.
\end{enumerate}

The rest of this paper is structured as follows. We first introduce some basic notions in Section \ref{sec:concept} and present our attention-based query answering model in Section \ref{sec:method}. In Section \ref{sec:exp}, we discuss the datasets we used to evaluate our model and present the evaluation results. We conclude our work in Section \ref{sec:con}.

\section{Basic Concepts} \label{sec:concept}
\label{subsec:concept}
Before  introducing our end-to-end attention-based logical query answering model, we outline some basic notions relevant to Conjunctive Graph Query models.




\vspace{-0.1cm}
\subsection{Conjunctive Graph Queries (CGQ)} 

In this work, a knowledge graph (KG) is a directed and labeled multi-relational graph $\kg = (\entset, \relset)$  where $\entset$ is a set of entities (nodes), $\relset$ is the set of relations (predicates, edges); furthermore let $\triset$ be a set of triples. A triple $T_{i} = (h_{i},r_{i},t_{i})$ or $\rel_i(h_i,t_i)$ in this sense consists of a head entity $h_{i}$ and a tail entity $t_{i}$ connected by some relation $r_{i}$ (predicate).\footnote{Note that in many knowledge graphs, a triple can include a datatype property as the relation where the tail is a literal. In line with related work \cite{wang2017knowledge,nickel2016review} we do not consider this kind of triples here. We will use head (h), relation (r), and tail(t) when discussing embeddings and subject (s), predicate (p), object (o) when discussing Semantic Web knowledge graphs to stay in line with the literature from both fields.}

\begin{definition}[Conjunctive Graph Query (CGQ)]
	\label{def:cgq}
	A query $\cgq \in Q(\kg)$ that can be written as follows:
	\begin{eqnarray*}
		& \cgq = \variable_{?}. \exists \variable_{1}, \variable_{2},..,\variable_{m}: \gp_{1} \land \gp_{2} \land ... \land \gp_{n} \\
		where &\; \gp_{i} = \rel(\ent_{k}, \variable_{l}), \variable_{l} \in \{\variable_{?},  \variable_{1}, \variable_{2},..,\variable_{m}\}, \ent_{k} \in \entset, \rel \in \relset \\
		or& \; \gp_{i} = \rel(\variable_{k}, \variable_{l}), \variable_{k}, \variable_{l} \in \{\variable_{?},  \variable_{1}, \variable_{2},..,\variable_{m}\}, k \neq l, \rel \in \relset
		\label{equ:cgq}
	\end{eqnarray*}
\end{definition}
Here $\variable_{?}$ denotes the target variable of the query which will be replaced with the answer entity, while $\variable_{1}, \variable_{2},..,\variable_{m}$ are existentially quantified bound variables. $\gp_{i}$ is a basic graph pattern in this CGQ. To ensure $\cgq$ is a valid CGQ, the dependence graph of $\cgq$ must be a \textit{directed acyclic graph (DAG)} \cite{hamilton2018embedding} in which the entities (anchor nodes) $\ent_{k}$ in $\cgq$ are the source nodes and the target variable $\variable_{?}$ is the unique sink node.

Figure \ref{fig:qg} shows an example CGQ which is equivalent to the SPARQL query in Listing \ref{q:exp}, where \textit{?Person} is an existentially quantified bound variable and \textit{?Disease} is the target variable. Note that for graph pattern $\rel(s, o)$ where subject $s$ is a variable and object $o$ is an entity, we can convert it into the form $\gp_{i} = \rel(\ent_{k}, \variable_{l})$ by using the inverse relation of the predicate $r$. In other words, we convert $\rel(s, o)$ to $\rel^{-1}(o, s)$. For example, In Figure \ref{fig:qg}, we use $AlmaMater^{-1}(UCLA, ?Person)$ to represent the graph pattern $AlmaMater(?Person, UCLA)$. The benefit of this inverse relation conversion is that we can construct CGQ where the dependence graph is a \textit{directed acyclic graph (DAG)} as shown in Figure \ref{fig:qg} .

Comparing  Definition \ref{def:cgq} with 
SPARQL, we can see several differences:
\begin{enumerate}
	\item Predicates in CGQs are assumed to be fixed while predicates in a SPARQL 1.1 basic graph pattern can also be variables \cite{mai2019relaxing}.
	\item CGQs only consider the conjunction of graph patterns while SPARQL 1.1 also contains other operations (UNION, OPTION, FILTER, LIMIT,
	etc.).
	\item CGQs require one variable as the answer denotation, which is in alignment with most question answering over knowledge graph literature \cite{berant2013semantic,liang2017neural}. In contrast, SPARQL 1.1 allows multiple variables as the returning variables. The unique answer variable property make it easier to evaluate the performance of different deep learning models on CGQs.
\end{enumerate}

\subsection{Geometric Operators in  Embedding Space} \label{subsec:operator}

Here we  describe two geometric operators - the projection operator and the intersection operator - in the entity embedding space, which were first introduced by Hamilton et al.~\cite{hamilton2018embedding}.

\begin{definition}[Geometric Projection Operator]
	\label{def:project}
	Given an embedding $\entemb_{i} \in \realnum^{\embdim}$ in the entity embedding space which can be either an embedding of a real entity $\ent_{i}$ or a computed embedding for an existentially quantified bound variable $\variable_{i}$ in a conjunctive query $\cgq$, and a relation $\rel$, the projection operator $\proj$ produces a new embedding $\entemb_{i}^{\prime} = \proj(\ent_{i}, \rel)$ where $\entemb_{i}^{\prime} \in \realnum^{\embdim}$. The projection operator is defined as follows:
	\begin{equation}
		\entemb_{i}^{\prime} = \proj(\ent_{i}, \rel) = \projmat_{\rel}\entemb_{i}
		\label{equ:proj}
	\end{equation}
	where $\projmat_{\rel} \in \realnum^{\embdim \times \embdim}$ is a trainable and relation-specific matrix for relation type $\rel$.
	The embedding $\entemb_{i}^{\prime} = \proj(\ent_{i}, \rel)$ denotes all entities that connect with entity $\ent_{i}$ or variable $\variable_{i}$ through relation $\rel$. If embedding $\entemb_{i}$ denotes entity $\ent_{i}$, then $\entemb_{i}^{\prime} = \proj(\entemb_{i}, \rel)$ denotes $\{\ent_{k} | \rel(\ent_{i}, \ent_{k}) \in \kg\}$. If embedding $\entemb_{i}$ denotes variable $\variable_{i}$, then $\entemb_{i}^{\prime} = \proj(\ent_{i}, \rel)$ denotes $\{\ent_{k} | \ent_{j} \in \variable_{i} \land \rel(\ent_{j}, \ent_{k}) \in \kg\}$.
\end{definition}

In short, $\entemb_{i}^{\prime} = \proj(\entemb_{i}, \rel)$ denotes the embedding of the relation $\rel$ specific neighboring set of entities. Different KG embedding models have different ways to represent the relation $\rel$. We can also use TransE's version ($\entemb_{i}^{\prime} = \entemb_{i} + \mathbf{r}$) or a diagonal matrix version ($\entemb_{i}^{\prime} = diag(\mathbf{r})\entemb_{i}$, where $diag(\mathbf{r})$ is a diagonal matrix parameterized by vector $\mathbf{r}$ in its diagonal axis). The bilinear version shown in Equation \ref{equ:proj} has the best performance in logic query answering because it is more flexible in capturing different characteristics of relation $\rel$ \cite{hamilton2018embedding}. 

As for the intersection operator, we first present the original version from Graph Query Embedding (GQE) \cite{hamilton2018embedding}, which will act as  baseline for our model.
\begin{definition}[Geometric Intersection Operator]
	\label{def:intergqe}
	Assume we are given a set of $\neisize$\ different input embeddings $\entemb_{1}^{\prime}$, $\entemb_{2}^{\prime}$, ..., $\entemb_{i}^{\prime}$,..., $\entemb_{\neisize}^{\prime}$ as the outputs from $\neisize$\ different geometric projection operations $\proj$ by following $\neisize$\ different relation $\rel_{j}$ paths. We require all $\entemb_{i}^{\prime}$ to have the same entity type. The geometric intersection operator outputs one embedding $\entemb^{\prime\prime}$ based on this set of embeddings which denotes the intersection of these different relation paths:
{\scriptsize 	\begin{equation}
	\entemb^{\prime\prime} = \inter_{\baseline}(\{\entemb_{1}^{\prime}, \entemb_{2}^{\prime}, ..., \entemb_{i}^{\prime}, ..., \entemb_{\neisize}^{\prime}\}) = \intermat_{\type 1}\interelemfun(\relu(\intermat_{\type 2} \;\entemb_{i}^{\prime}), \forall i \in \{1,2,..,\neisize\})
	\label{equ:intergqe}
	\end{equation}}
	where $\intermat_{\type 1}, \intermat_{\type 2} \in \realnum^{\embdim \times \embdim}$ are trainable entity type $\type$ specific matrices. $\interelemfun()$ is a symmetric vector function (e.g., an element-wise mean or minimum of a set of vectors) which is permutation invariant on the order of its inputs \cite{zaheer2017deep}. 
	As $\entemb_{1}^{\prime}$, $\entemb_{2}^{\prime}$, ..., $\entemb_{i}^{\prime}$,..., $\entemb_{\neisize}^{\prime}$ represent the embeddings of the neighboring set of entities, $\entemb^{\prime\prime} = \inter_{\baseline}(\{\entemb_{1}^{\prime}, \entemb_{2}^{\prime}, ..., \entemb_{i}^{\prime}, ..., \entemb_{\neisize}^{\prime}\})$ is interpreted as the intersection of these sets.
\end{definition}
\subsection{Entity Embedding Initialization} \label{subsec:embed}

Generally speaking, any (knowledge) graph embedding model can be used to initialize entity embeddings. In this work, we adopt the simple ``bag-of-features'' approach. We assume each entity $\ent_{i}$ will have an entity type $\type = \typefun(e_{i})$, e.g. \texttt{Place}, \texttt{Agent}. The entity embedding lookup is shown below:

\begin{equation}
	\entemb_{i} = \dfrac{\embmat_{\type}\onehotvec_{i}}{\parallel\embmat_{\type}\onehotvec_{i}\parallel_{L2}}
\end{equation}

$\embmat_{\type} \in \realnum^{\embdim \times \numentpertype}$ is the type-specific embedding matrices for all entities with type $\type = \typefun(e_{i})$ which can be initialized using a normal embedding matrix normalization method. The $\onehotvec_{i}$ is a binary feature vector such as a one-hot vector which uniquely identifies entity $\ent_{i}$ among all entities with the same entity type $\type$. The $\parallel\cdot\parallel_{L2}$ indicates the $L2$-norm. The reason why we use type-specific embedding matrices rather than one embedding matrix for all entities as \cite{nickel2012factorizing,bordes2013translating,socher2013reasoning,yang2014embedding,lin2015learning,ji2015knowledge,nickel2016holographic} did is that recent node embedding work \cite{hamilton2017representation,hamilton2018embedding} show that most of the information contained in the node embeddings is type-specific information. Using type-specific entity embedding matrices explicitly handles this information. Note that in many KGs such as DBpedia one entity may have multiple types. We handle this by computing the common super class of these types (see Sec. \ref{sec:exp}).

\section{Method} \label{sec:method}

Next, we discuss the difference between our model and GQE \cite{hamilton2018embedding}. Our geometric operators (1) use an attention mechanism to account for the fact that different paths have different embedding prediction abilities with respect to the center entity embedding and (2) can be applied to two training phases -- training on the original KG and training with sampled logic query-answer pairs.

\subsection{Attention-based Geometric Projection Operator}
	\label{AGPO}

Since the permutation invariant function $\interelemfun()$ directly operates on the set $\{ReLU(\intermat_{\type 2} \;\entemb_{i}^{\prime}) | \forall i \in \{1,2,..,\neisize\} \}$, Equation \ref{equ:intergqe} assumes that each $\entemb_{i}^{\prime}$ (relation path) has an equal contribution to the final intersection embedding $\entemb^{\prime\prime}$. This is not necessarily the case in real settings as we have discussed in Section \ref{sec:intro}. Graph Attention Network (GAT) \cite{velivckovic2017graph} has shown that using an attention mechanism on graph-structured data to capture the unequal contribution of the neighboring nodes to the center node yields better result than a simple element-wise mean or minimum approaches. By following the attention idea of GAT, we propose an attention-based geometric intersection operator. 

	Assume we are given the same input as Definition \ref{def:intergqe}, a set of $\neisize$ different input embeddings $\entemb_{1}^{\prime}$, $\entemb_{2}^{\prime}$, ..., $\entemb_{i}^{\prime}$,..., $\entemb_{\neisize}^{\prime}$. The geometric intersection operator contains two layers: a multi-head attention layer and a feed forward neural network layer.
	
	\subsubsection{The multi-head attention layer}
	The initial intersection embedding $\entemb_{init}^{\prime\prime}$ is computed as:
	\begin{equation}
	\entemb_{init}^{\prime\prime} = \interelemfun(\entemb_{i}^{\prime}, \forall i \in \{1,2,..,\neisize\})
	\label{equ:interinit}
	\end{equation}
	
	Then the attention coefficient for each $\entemb_{i}^{\prime}$ in the $\attnidx^{th}$ attention head is
	\begin{equation}
	\attnscr_{i\attnidx} = \attnfun_{\attnidx}(\entemb_{init}^{\prime\prime}, \entemb_{i}^{\prime})
	 = \dfrac{exp(\leakyrelu(\attnvec_{\type\attnidx}^{T}[\entemb_{init}^{\prime\prime};\entemb_{i}^{\prime}]))}{\sum_{j=1}^{\neisize} exp(\leakyrelu(\attnvec_{\type\attnidx}^{T}[\entemb_{init}^{\prime\prime};\entemb_{j}^{\prime}]))}
	\label{equ:atten}
	\end{equation}
	where $\cdot^{T}$ represents transposition, $[;]$  vector concatenation, and $\attnvec_{\type\attnidx} \in \realnum^{\embdim \times 2} $ is the type-specific trainable attention vector for $\attnidx^{th}$ attention head. Following the advice on avoiding spurious weights~\cite{velivckovic2017graph}, we use \textit{LeakyReLu} here.
	
	The attention weighted embedding $\entemb_{attn}^{\prime\prime}$ is computed as the weighted average of different input embeddings while weights are automatically learned by the multi-head attention mechanism. Here, $\attnact()$ is the sigmoid activation function and $\attnsize$ is the number of attention heads.
	\begin{equation}
	\entemb_{attn}^{\prime\prime} = \attnact(\dfrac{1}{\attnsize} \sum_{\attnidx=1}^{\attnsize} \sum_{i=1}^{\neisize} \attnscr_{i\attnidx}\entemb_{i}^{\prime}) 
	\label{equ:multiatten}
	\end{equation}
	Furthermore, we add a residual connection \cite{he2016deep} of $\entemb_{attn}^{\prime\prime}$, followed by layer normalization \cite{ba2016layer} (Add \& Norm).
	\begin{equation}
	\entemb_{ln1}^{\prime\prime} = \lynorm_{1}(\entemb_{attn}^{\prime\prime} + \entemb_{init}^{\prime\prime})
	\label{equ:layernorm1}
	\end{equation}
	
	\subsubsection{The second layer} It is a normal feed forward neural network layer
	followed by the ``Add \& Norm'' as shown in Equation \ref{equ:layernorm2}.
{\footnotesize 	\begin{equation}
	\entemb^{\prime\prime} = \inter_{\model}(\{\entemb_{1}^{\prime}, \entemb_{2}^{\prime}, ..., \entemb_{i}^{\prime}, ..., \entemb_{\neisize}^{\prime}\}) = \lynorm_{2}(\intermat_{\type}\entemb_{ln1}^{\prime\prime} + \interbia_{\type} + \entemb_{ln1}^{\prime\prime})
	\label{equ:layernorm2}
	\end{equation}}
	where $\intermat_{\type} \in \realnum^{\embdim \times \embdim}$ and $\interbia_{\type} \in \realnum^{\embdim}$ are trainable entity type $\type$ specific weight matrix and bias vector, respectively, in a feed forward neural network.

\begin{figure}[]
	\centering
	\includegraphics[width=1.0\columnwidth]{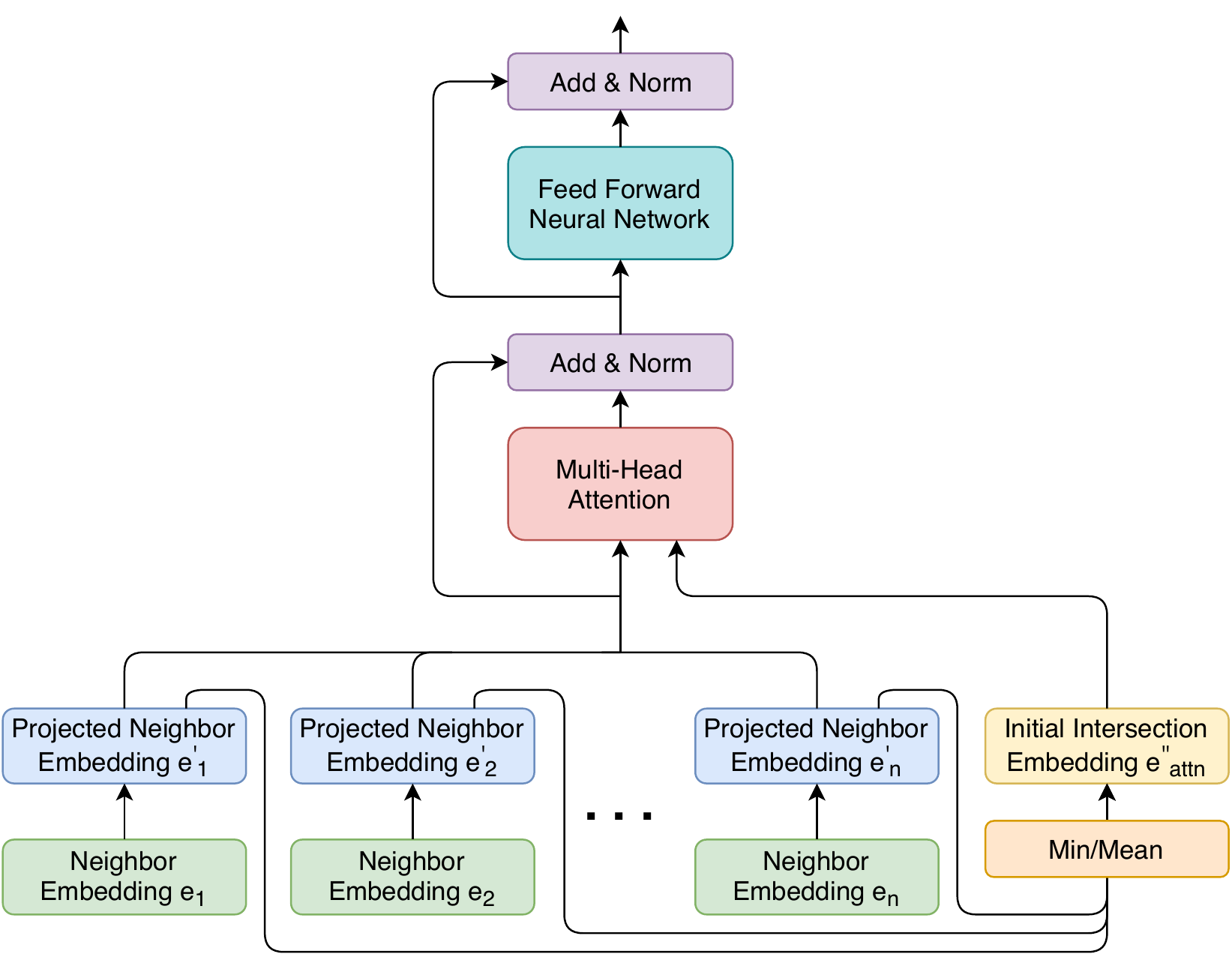}
	\caption{The attention-based geometric intersection operator - model architecture} \label{fig:atten}
\end{figure}

Figure \ref{fig:atten} illustrates the model architecture of our attention-based geometric intersection operator. The light green boxes at the bottom indicate $\neisize$ embeddings $\entemb_{1}$, $\entemb_{2}$,...,$\entemb_{i}$,...,$\entemb_{\neisize}$, which are projected by the geometric projection operators. The output embeddings $\entemb_{1}^{\prime}$, $\entemb_{2}^{\prime}$, ..., $\entemb_{i}^{\prime}$,..., $\entemb_{\neisize}^{\prime}$ are the $\neisize$\ input embeddings of our intersection operator. The initial intersection embedding $\entemb_{init}^{\prime\prime}$ is computed based on these input embeddings as shown in Equation \ref{equ:interinit}. Next, $\entemb_{init}^{\prime\prime}$ and $\entemb_{1}^{\prime}$, $\entemb_{2}^{\prime}$, ..., $\entemb_{i}^{\prime}$,..., $\entemb_{\neisize}^{\prime}$ are fed into the multi-head attention layer followed by the feed forward neural network layer. This two-layer architecture is inspired by Transformer \cite{vaswani2017attention}.

The multi-head attention mechanism shown in Equation \ref{equ:interinit}, \ref{equ:atten}, and \ref{equ:multiatten} is similar to those used in Graph Attention Network (GAT) \cite{velivckovic2017graph}. The major difference is the way we compute the initial intersection embedding $\entemb_{init}^{\prime\prime}$ in Equation \ref{equ:interinit}. In the graph neural network context, the attention function can be interpreted as mapping \textit{the center node embedding} and \textit{a set of neighboring node embeddings} to an output embedding. In GAT, the model directly operates on the local graph structure by applying one or multiple convolution operations over the 1-degree neighboring nodes of the center node. In order to compute the attention score for each neighboring node embedding, each of the neighboring node embedding is compared with the embedding of the center node for attention score computation. Here, the center node embedding is known in advance. 

However, in our case, since we want to train our model directly on the logical query-answer pairs (\textbf{query-answer pair training phase}), the final intersection embedding $\entemb^{\prime\prime}$ might denote the variable in a conjunctive graph query $q$ whose embedding is unknown. For example, in Figure \ref{fig:qg}, we can obtain two embeddings $\mathbf{p_{1}}$ and $\mathbf{p_{2}}$ for variable \textit{?Person} by following two different triple path $T_{1}$ and $T_{2}$. In this case, the input embeddings for our intersection operator are $\mathbf{p_{1}}$ and $\mathbf{p_{2}}$. The \textit{center node embedding} here is the true embedding for variable \textit{?Person} which is unknown. Equation \ref{equ:interinit} is used to compute an initial embedding for the center node, the variable \textit{?Person}, in order to compute the attention score for each input embedding.

Note that these two intersection operators in Definition \ref{def:intergqe} and Section \ref{AGPO} can also be \textit{directly applied to the local knowledge graph structure} as R-GCN \cite{schlichtkrull2018modeling} does (\textbf{original KG training phase}). The output embedding $\entemb^{\prime\prime}$ can be used as the new embedding for the center entity which is computed by a convolution operation over its 1-degree neighboring entity-relation pairs. In this KG training phase, although the center node embedding is known in advance, in order to make our model applicable to both of these two training phases, we still use the initial intersection embedding idea. Note that the initial intersection embedding computing step (see Equation \ref{equ:interinit}) solves the problem of the previous attention mechanism where the center node embedding is a prerequisite for attention score computing. This makes our graph attention mechanism applicable to both logic query answering and KG embedding training. As far as we know, it is the first graph attention mechanism applied on both tasks.

\subsection{Model Training} \label{subsec:train}

The projection operator and intersection operator constitute our attention-based logical query answering model.
As for the model training, it has two training phases: the original KG training phase and the query-answer pair training phase.

\subsubsection{Original KG Training Phase} \label{subsubsec:kgtrain}
In original KG training phase, we train those two geometric operators based on the local KG structure. Given a KG $\kg = \langle \entset, \relset \rangle$, for every entity $\ent_{i} \in \entset$, we use the geometric projection and intersection operator to compute a new embedding $\entemb_{i}^{\prime\prime}$ for entity  $\ent_{i}$ given its 1-degree neighborhood $\nei(\ent_{i}) = \{(\rel_{ui}, \ent_{ui}) | \rel_{ui}(\ent_{ui}, \ent_{i}) \in \kg\} \cup \{(\rel_{oi}^{-1}, \ent_{oi}) | \rel_{oi}(\ent_{i}, \ent_{oi}) \in \kg\}$ which is a sampled set of neighboring entity-relation pairs with size $\neisize$. Here, $\inter()$ indicates either $\inter_{\baseline}()$ (baseline model) or $\inter_{\model}()$ (proposed model).
\begin{equation}
\entemb_{i}^{\prime\prime} = \kgtembed(\ent_{i}) = \inter(\{\proj(\ent_{ci}, \rel_{ci}) | (\rel_{ci}, \ent_{ci}) \in \nei(\ent_{i})\})
\label{equ:kgtrain}
\end{equation}

Let $\entemb_{i}$ indicates the true entity embedding for $\ent_{i}$ and $\entemb_{i}^{-}$ indicates the embedding of a negative sample $\ent_{i}^{-} \in \negsamp(\ent_{i})$, where $\negsamp(\ent_{i})$ is the negative sample set for $\ent_{i}$. The loss function for this KG training phase is a max-margin loss:
{\footnotesize \begin{equation}
\kgtloss = \sum_{\ent_{i} \in \entset} \sum_{\ent_{i}^{-} \in \negsamp(\ent_{i})} max(0, \Delta - \cosine(\kgtembed(\ent_{i}), \entemb_{i}) + \cosine(\kgtembed(\ent_{i}), \entemb_{i}^{-}))
\label{equ:kgloss}
\end{equation}}

Here $\Delta$ is margin and $\cosine()$ denote the cosine similarity function:
\begin{equation}
\cosine(\queryembed, \answerembed) = \dfrac{\queryembed \cdot \answerembed}{\parallel \queryembed \parallel  \parallel \answerembed \parallel}
\label{equ:cos}
\end{equation}

\subsubsection{Logical Query-Answer Pair Training Phase} \label{subsubsec:qatrain}

In this training phase, we first sample $\qasamplesize$\ different conjunctive graph query (logical query)-answer pairs $\qaset = \{(\query_{i}, \answer_{i})\}$ from the original KG by sampling entities at each node in the conjunctive query structure according to the topological order (See Hamilton et al. \cite{hamilton2018embedding}). Then for each conjunctive graph query $\query_{i}$ with one or multiple anchor nodes $\{\ent_{i1},\ent_{i2},..,\ent_{in}\}$, we compute the embedding for its target variable node $\variable_{i?}$, denote as $\queryembed_{i}$, based on two proposed geometric operators (See Algorithm 1 in Hamilton et al. \cite{hamilton2018embedding} for a detailed explanation). We denote the embedding for the correct answer entity as $\answerembed_{i}$ and the embedding for the  negative answer as $\answerembed_{i}^{-}$ where $\answer_{i}^{-} \in \negsamp(\query_{i}, \answer_{i})$. The loss function for this query-answering pair train phase is:
{\footnotesize \begin{equation}
\qatloss = \sum_{(\query_{i}, \answer_{i}) \in \qaset} \sum_{\answer_{i}^{-} \in \negsamp(\query_{i}, \answer_{i})} max(0, \Delta - \cosine(\queryembed_{i}, \answerembed_{i}) + \cosine(\queryembed_{i}, \answerembed_{i}^{-}))
\label{equ:qaloss}
\end{equation}}


\subsubsection{Negative Sampling} \label{subsubsec:neg_sample}
As for negative sampling method, we adopt two methods: 1) \textit{negative sampling}: $\negsamp(\ent_{i})$ is a fixed-size set of entities which have the same entity type as $\ent_{i}$ except $\ent_{i}$ itself; 2) \textit{hard negative sampling}: $\negsamp(\ent_{i})$ is a fixed-size set of entities which satisfy some of the entity-relation pairs in $\nei(\ent_{i})$ but not all of them.

\subsubsection{Full Model Training}
The loss function for the whole model training is the combination of these two training phases:

\begin{equation}
\loss = \kgtloss + \qatloss
\label{equ:loss}
\end{equation}

While Hamilton et al. \cite{hamilton2018embedding} trains the model only using logical query-answer pair training phase and Equation \ref{equ:qaloss} as the loss function. We generalize their approach by adding the KG training phase to better incorporate the KG structure into the training.

\section{Experiment} \label{sec:exp}

We carried out empirical study following the experiment protocol of Hamilton et al. \cite{hamilton2018embedding}. 
To properly test all models' ability to reason with larger knowledge graph of many relations, we constructed two datasets from publicly available \textit{DBpedia} and \textit{Wikidata}.

{\small \begin{table*}
	\caption{Statistics for Bio, \db\ and \wiki~ (Section~\ref{subsec:data}).
	``NUM/QT'' indicates the number of QA pairs per query type.}
	\label{tab:data}
	\centering
	
\begin{tabular}{l|rrr|rrr|rrr}
\toprule
                              & \multicolumn{3}{c|}{Bio}          & \multicolumn{3}{c|}{DB18}         & \multicolumn{3}{c}{WikiGeo19}    \\
                              & Training & Validation & Testing  & Training & Validation & Testing  & Training & Validation & Testing  \\ \hline
\# of Triples                 & 3,258,473  & 20,114      & 181,028   & 122,243   & 1,358       & 12,224    & 170,409   & 1,893       & 17,041    \\
\# of Entities                & 162,622   & -          & -        & 21,953    & -          & -        & 18,782    & -          & -        \\
\# of Relations               & 46       & -          & -        & 175      & -          & -        & 192      & -          & -        \\ \hline
\# of Sampled 2-edge QA Pairs & 1M  & 1k/QT    & 10k/QT & 1M  & 1k/QT    & 10k/QT & 1M  & 1k/QT    & 10k/QT \\
\# of Sampled 3-edge QA Pairs & 1M  & 1k/QT    & 10k/QT & 1M  & 1k/QT    & 10k/QT & 1M  & 1k/QT    & 10k/QT \\
\bottomrule
\end{tabular}
\vspace{-0.3cm}
\end{table*}}

\subsection{Datasets} \label{subsec:data}

Hamilton et al. \cite{hamilton2018embedding} conducted logic query answering evaluation with Biological interaction and Reddits videogame datasets\footnote{\url{https://github.com/williamleif/graphqembed}}.
However, the reddit dataset is not made publicly available. 
The Bio interaction dataset has some issue of their logic query generation process\footnote{Hamilton et al. \cite{hamilton2018embedding} sample the training queries from the whole KG rather than the training KG, which makes all the triples in the KG known to the model and makes the tasks simpler than realistic test situations.}. Therefore, we regenerate the train/valid/test queries from the Bio KG. Furthermore, the Bio interaction dataset has only 46 relation types which is very simple compared to many widely used knowledge graphs such as \textit{DBpedia} and \textit{Wikidata}. 
Therefore we construct two more datasets (\db and \wiki) with
larger graphs and more relations based on \textit{DBpedia} and \textit{Wikidata}
\footnote{The code and both datasets are available at \url{https://github.com/gengchenmai/Attention_GraphQA}.}. 

Both datasets are constructed in a similar manner as \cite{hamilton2018embedding}:
\begin{enumerate}
	\item First collect a set of \textit{seed entities};
	\item Use these seed entities to get their 1-degree and 2-degree object property triples;
	\item Delete the entities and their associated triples with node degree less than a threshold \datathreshold; 
	\item Split the triple set into training, validation, and testing set and make sure that every entity and relation in the validation and testing dataset will appear in training dataset. The training/validation/testing split ratio is 90\%/1\%/9\%;
	\item Sample the training queries from the training KG\footnote{We modify the query generation code provided by Hamilton et al. \cite{hamilton2018embedding}}.
\end{enumerate}

For \textbf{\db} 
the seed entities are all geographic entities directly linked to \path{dbr:California} via \path{dbo:isPartOf} with type (\path{rdf:type}) \path{dbo:City}. There are 462 seed entities in total. In Step 2, we filter out triples with no \path{dbo:} prefixed properties. The threshold \datathreshold\ is set up to be 10.  
For \textbf{\wiki} 
the seed entities are the largest cities in each state of the United States\footnote{\url{https://www.infoplease.com/us/states/state-capitals-and-largest-cities}}. 
The threshold \datathreshold\ is 20 which is a relatively small value compare to  \datathreshold=100 for the widely used FB15K and WN18 dataset.
Statistic for these 3 datasets are shown in Table \ref{tab:data}. Given that the widely used KG completion dataset FB15K and WN18 have 15K and 41K triples, \db~and \wiki~are rather large in size (120K and 170K triples). Note that for each triple $r(s,o)$ in  training/validation/testing dataset, we also add its inverse relation $r^{-1}(o,s)$ to the corresponding dataset and \textit{the geometric projection operator will learn two separated projection matrices $\projmat_{\rel}$ $\projmat_{\rel^{-1}}$ for each relation}. The training triples constitute the training KG. Note that both \baseline\ and \model\ require to know the unique type for each entity. However, entities in \textit{DBpedia} and \textit{Wikidata} have multiple types (\path{rdf:type}). As for \db, we utilize the level-1 classes in \textit{DBpedia} ontology and classify each entity to these level-1 classes based on the \path{rdfs:subClassOf} relationships. For \wiki, we simply annotate each entity with class \textit{Entity}.

\subsection{Training Details}
\label{subsec:train_details}
As we discussed in Section \ref{subsec:train}, we train our \model\ model based on two training phases. In the original KG training phase, we adopt an minibatch training strategy. In order to speed up the model training process, we sample the neighborhood for each entity with different neighborhood sample size ($\neisize = 4,5,6,7$) in the training KG beforehand. We split these sampled node-neighborhood pairs by their neighborhood sample size $\neisize$ in order to do minibatch training. 

As for the logical query-answer pair training phase, we adopt the same query-answer pair sampling strategy as Hamilton et al. \cite{hamilton2018embedding}. We consider 7 different conjunctive graph query structures shown in Figure \ref{fig:dbauc}. As for the 4 query structures with intersection pattern, we apply hard negative sampling (see Section \ref{subsubsec:neg_sample}) 
and indicate them as 4 separate query types. In total, we have 11 query types. All training (validation/testing) triples are utilized as 1-edge conjunctive graph queries for model training (evaluation). As for 2-edge and 3-edge queries, the number for sampled queries for training/validation/testing are shown in Table \ref{tab:data}. Note that all training queries are sampled from the training KG. All validation and testing queries are sampled from the whole KG and we make sure these queries cannot be directly answered based on the training KG (unanswerable queries \cite{mai2019relaxing}). To ensure these queries are truly unanswerable, the matched triple patterns of these queries should contain at least one triple in the testing/validation triple set. 

\subsection{Baselines}
We use 6 different models as baselines: two models with the billinear projection operator $\entemb_{i}^{\prime} = \proj(\ent_{i}, \rel)$ and the element-wise mean or min as the simple intersection operator: \textbf{Billinear[mean\_simple]}, \textbf{Billinear[min\_simple]}; two models with the TransE based projection operator and the GQE version of geometric intersection operator: \textbf{TransE[mean]}, \textbf{TransE[min]}; and two GQE models \cite{hamilton2018embedding}: \textbf{\baseline[mean]}, \textbf{\baseline[min]}. Since $\interelemfun()$ can be element-wise mean or min, we differentiate them using \textbf{[mean]} and \textbf{[min]}. Note that all of these 6 baseline models only use the logical query-answer pair training phase (see Section \ref{subsubsec:qatrain}) to train the model.  As for model with billinear projection operator, based on multiple experiments, we find that the model with element-wise min consistently outperforms the model with element-wise mean. Hence for our  model, we use element-wise min for $\interelemfun()$.

\subsection{Results} \label{subsec:res}
We first test the effect of the origin KG training on the model performance without the attention mechanism called 
\textbf{\baseline+KG{[}min{]}}
here. Then we test the models with different numbers of attention heads with the added original KG training phase which are indicated as 
\cga{+x}, where \textbf{x} represents the number of attention heads (can be $1, 4, 8$).

\begin{table*}[t]
	\caption{
	Macro-average AUC and APR over test queries with different DAG structures are used to evaluate the performance. \textit{All} and \textit{H-Neg.} denote macro-averaged across all query types and query types with hard negative sampling (see Section \ref{subsubsec:neg_sample}). 
	}
	\label{tab:eval}
	\centering
	\vspace*{-0.2cm}
	\small {
	\begin{tabular}{c|cc|cc|cc|cc|cc|cc}
	\toprule
Dataset                            & \multicolumn{4}{c|}{Bio}                                         & \multicolumn{4}{c|}{DB18}                                          & \multicolumn{4}{c}{WikiGeo19}                                     \\ \hline
Metric                      & \multicolumn{2}{c|}{AUC}         & \multicolumn{2}{c|}{APR}       & \multicolumn{2}{c|}{AUC}         & \multicolumn{2}{c|}{APR}         & \multicolumn{2}{c|}{AUC}         & \multicolumn{2}{c}{APR}         \\
                            & All            & H-Neg      & All           & H-Neg     & All            & H-Neg      & All            & H-Neg      & All            & H-Neg      & All            & H-Neg      \\ \hline
Billinear{[}mean\_simple{]} & 81.65          & 67.26          & 82.39         & 70.07         & 82.85          & 64.44          & 85.57          & 71.72          & 81.82          & 60.64          & 82.35          & 64.22          \\
Billinear{[}min\_simple{]}  & 82.52          & 69.06          & 83.65         & 72.7          & 82.96          & 64.66          & 86.22          & 73.19          & 82.08          & 61.25          & 82.84          & 64.99          \\
TransE{[}mean{]}            & 80.64          & 73.75          & 81.37         & 76.09         & 82.76          & 65.74          & 85.45          & 72.11          & 80.56          & 65.21          & 81.98          & 68.12          \\
TransE{[}min{]}             & 80.26          & 72.71          & 80.97         & 75.03         & 81.77          & 63.95          & 84.42          & 70.06          & 80.22          & 64.57          & 81.51          & 67.14          \\
\baseline{[}mean{]}               & 83.4           & 71.76          & 83.82         & 73.41         & 83.38          & 65.82          & 85.63          & 71.77          & 83.1           & 63.51          & 83.81          & 66.98          \\
\baseline{[}min{]}                & 83.12          & 70.88          & 83.59         & 73.38         & 83.47          & 66.25          & 86.09          & 73.19          & 83.26          & 63.8           & 84.3           & 67.95          \\ \hline
\baseline+KG{[}min{]}             & 83.69          & 72.23          & 84.07         & 74.3          & 84.23          & 68.06          & 86.32          & 73.49          & 83.66          & 64.48          & 84.73          & 68.51          \\ \hline
\cga{+1}           & 84.57          & 74.87          & 85.18         & 77.11         & 84.31          & 67.72          & 87.06          & 74.94          & 83.91          & 64.83          & 85.03          & 69             \\
\cga{+4}          & \textbf{85.13} & \textbf{76.12} & 85.46         & \textbf{77.8} & 84.46          & 67.88          & 87.05          & 74.66          & 83.96          & 64.96          & 85.36          & 69.64          \\
\cga{+8}           & 85.04          & 76.05          & \textbf{85.5} & 77.76         & \textbf{84.67} & \textbf{68.56} & \textbf{87.29} & \textbf{75.23} & \textbf{84.15} & \textbf{65.23} & \textbf{85.69} & \textbf{70.28} \\ \hline
Relative $\Delta$ over GQE& 2.31           & \textbf{7.29}  & 2.28          & \textbf{5.97} & 1.44           & \textbf{3.49}  & 1.39           & \textbf{2.79}  & 1.07           & \textbf{2.24}  & 1.65           & \textbf{3.43} \\
\bottomrule
\end{tabular}
	}
\vspace*{-0.2cm}
\end{table*}

\begin{figure*}[h!]
	\centering
	\begin{subfigure}[b]{1.0\columnwidth}
		\centering
		\includegraphics[width=\textwidth]{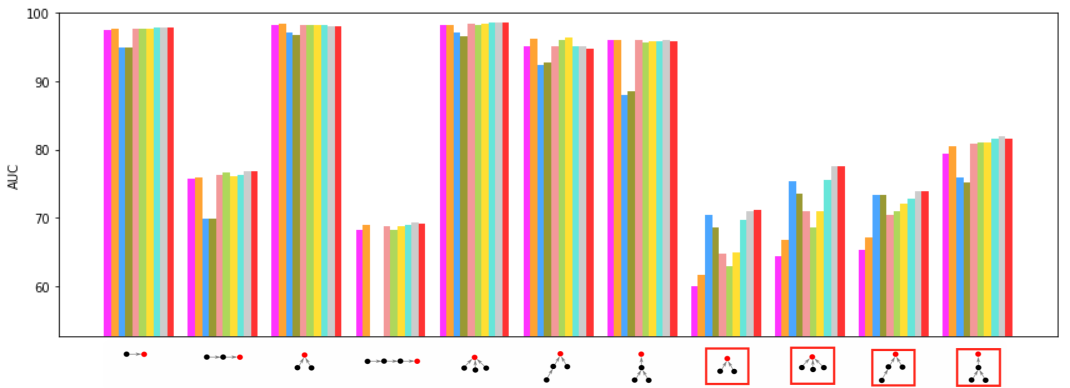}
		\caption[]%
		{{\small AUC for Bio}}    
		\label{fig:bioauc}
	\end{subfigure}
	\hfill
	\begin{subfigure}[b]{1.0\columnwidth}  
		\centering 
		\includegraphics[width=\textwidth]{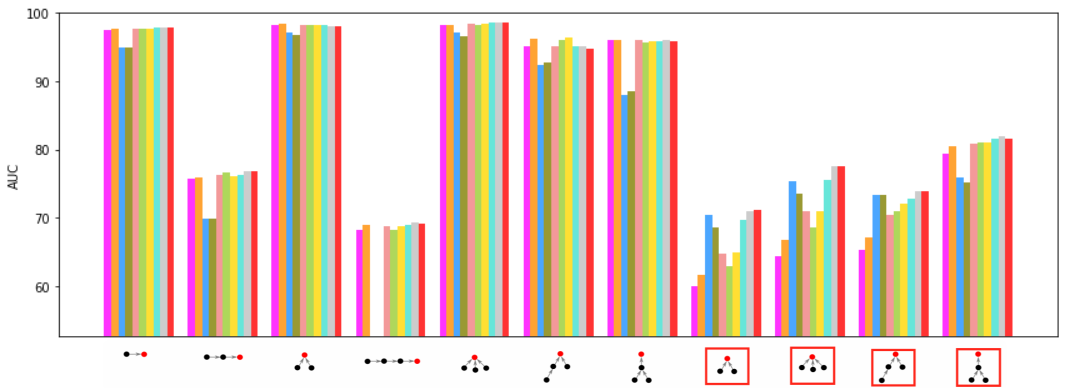}
		\caption[]%
		{{\small APR for Bio}}    
		\label{fig:bioapr}
	\end{subfigure}
	\vskip\baselineskip
	\begin{subfigure}[b]{1.0\columnwidth}
		\centering
		\includegraphics[width=\textwidth]{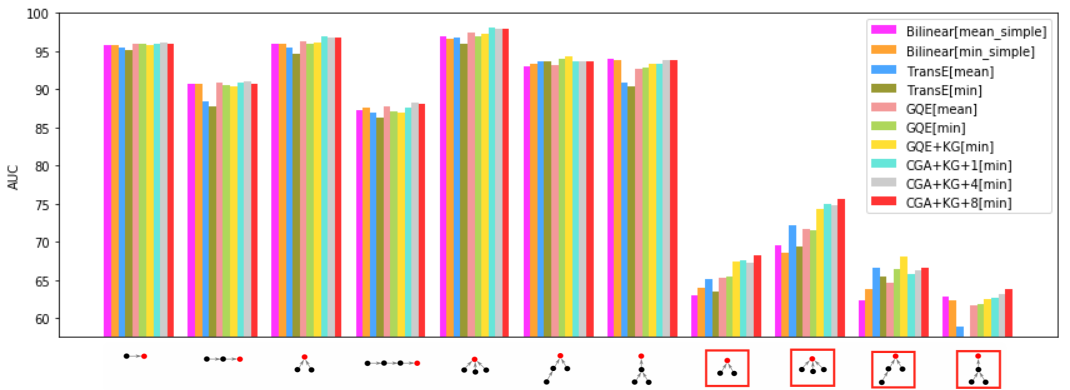}
		\caption[]%
		{{\small AUC for \db}}    
		\label{fig:dbauc}
	\end{subfigure}
	\hfill
	\begin{subfigure}[b]{1.0\columnwidth}  
		\centering 
		\includegraphics[width=\textwidth]{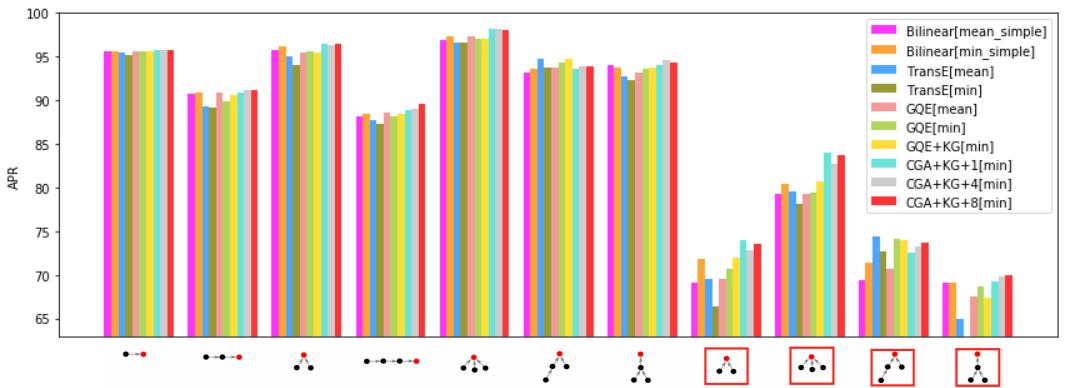}
		\caption[]%
		{{\small APR for \db}}    
		\label{fig:dbapr}
	\end{subfigure}
	\vskip\baselineskip
	\begin{subfigure}[b]{1.0\columnwidth}   
		\centering 
		\includegraphics[width=\textwidth]{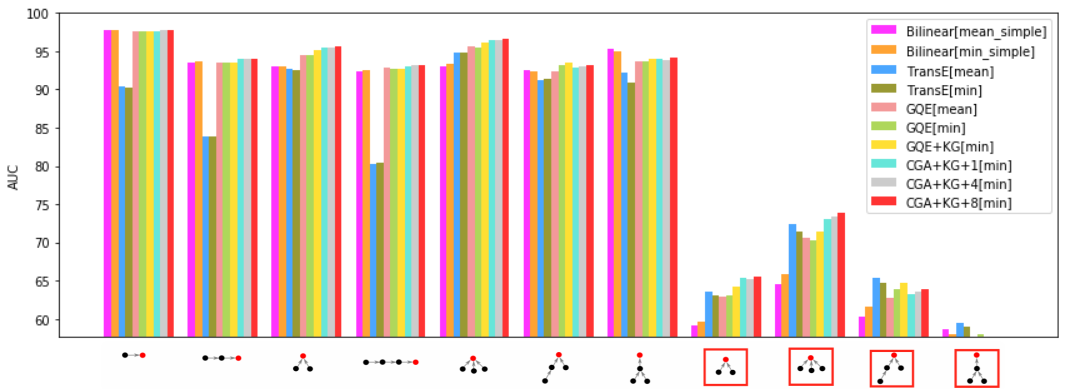}
		\caption[]%
		{{\small AUC for \wiki}}    
		\label{fig:wkauc}
	\end{subfigure}
	\quad
	\begin{subfigure}[b]{1.0\columnwidth}   
		\centering 
		\includegraphics[width=\textwidth]{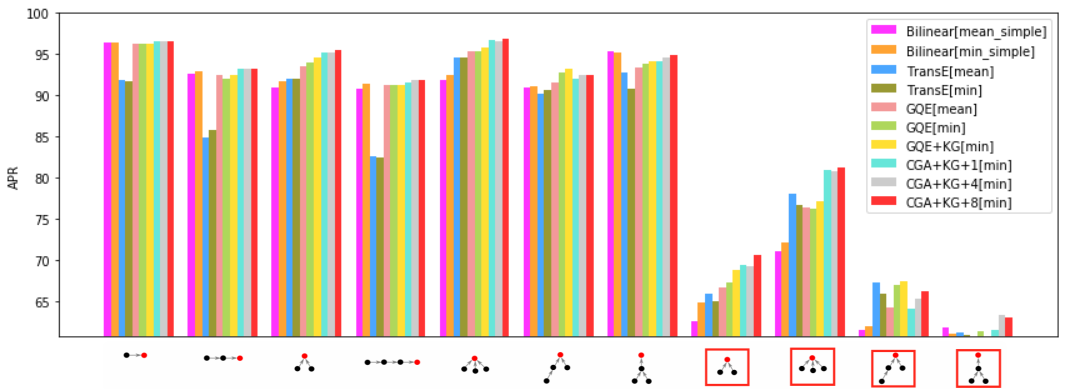}
		\caption[]%
		{{\small APR for \wiki}}    
		\label{fig:wkapr}
	\end{subfigure}
	\caption{Individual AUC and APR scores for different models per query type. Red boxes denote query types with hard negative sampling strategy} 
	\label{fig:eval_q}
	\vspace*{-0.55cm}
\end{figure*}

Table \ref{tab:eval} shows the evaluation results of the baseline models as well as different variations of our models on the test queries. We use the ROC AUC score and average percentile rank (APR) as two evaluation metrics. All evaluation results are macro-averaged across queries with different DAG structures (Figure \ref{fig:dbauc}). 

\begin{enumerate}
	\item All 3 variations of \model\ 
	consistently outperform 
	baseline models with fair margins which indicates the effectiveness of contextual attention.
The advantage 
is more obvious in query types with hard negative queries. 
	\item Comparing \textbf{\baseline+KG{[}min{]}} with other baseline models we can see that adding the original KG training phase in the model training process  improves the model performance. This shows that the structure information of the original KG is very critical for knowledge graph embedding model training even if the task is not link prediction.
	\item Adding the attention mechanism further improves the model performance. 
	This indicates the importance of
    considering the unequal contribution of the neighboring nodes to the center node embedding prediction.
	\item Multi-head attention models outperforms single-head models which is consistent with the result from GAT \cite{velivckovic2017graph}.
	\item Theoretically, $\inter_{\baseline}()$ has $2Ld^2=Ld(2d)$ learnable parameters while $\inter_{\model}()$ has $Ld^2+2KLd+Ld=Ld(d+2K+1)$ parameters where $L$ is the total number of entity types in a KG. Since usually $K \ll d$, \textit{our model has fewer parameters than GQE while achieves better performance.}
	\item \model~{} shows strong advantages over baseline models especially on query types with hard negative sampling (e.g., 7.3\% relative AUC improvement over GQE on Bio dataset\footnote{Note that since we regenerate queries for Bio dataset, the GQE performance is lower than the reported performance in Hamilton et al. \cite{hamilton2018embedding} which is understandable.}).
\end{enumerate}

All models shown in Table \ref{tab:eval} are implemented in PyTorch based on the official code\footnote{\url{https://github.com/williamleif/graphqembed}} of Hamilton et al. \cite{hamilton2018embedding}. 
The hyper-parameters for the baseline models GQE are tuned using grid search and the best ones are selected. Then we follow the practice of Hamilton et al. \cite{hamilton2018embedding} and used the same hyper-parameter settings for our CGA models:
128 for embedding dimension $\embdim$, 0.001 for learning rate, 512 for batch size. We use Adam optimizer for model optimization.

The overall delta of \model\ over \baseline\ reported in Tab. \ref{tab:eval} is similar in magnitude to the delta over baseline reported in Hamilton et al. \cite{hamilton2018embedding}. This is because \model\ will significantly outperform \baseline\ in query types with intersection structures, e.g., the 9th query type in Fig. \ref{fig:dbauc}, but perform on par in query types which do not contain intersection, e.g. the 1st query type in Fig. \ref{fig:dbauc}. Macro-average computation over all query types makes the improvement less obvious. In order to compare the performance of different models on different query structures (different query types), we show the individual AUC and APR scores on each query type in three datasets for all models (See Figure \ref{fig:bioauc}, \ref{fig:bioapr}, \ref{fig:dbauc}, \ref{fig:dbapr}, \ref{fig:wkauc}, and \ref{fig:wkapr}). To highlight the difference, we subtract the minimum score from the other scores in each figure. We can see that our model consistently outperforms the baseline models in almost all query types on all datasets except for the sixth and tenth query type (see Figure \ref{fig:eval_q}) which correspond to the same query structure \textit{3-inter\_chain}. In both these two query types, \textbf{\baseline+KG{[}min{]}} has the best performance. The advantage of our attention-based models is more obvious for query types with hard negative sampling strategy. 
For example, as for the 9th query type (\text{Hard-3-inter}) in Fig. \ref{fig:dbapr}, \cga{+8} has \textbf{5.8\%} and \textbf{6.5\%} relative APR improvement (\textbf{5.9\%} and \textbf{5.1\%} relative AUC improvement) over \baseline{[}min{]} on \db~{} and \wiki. Note that this query type has the largest number of neighboring nodes (3 nodes) which shows that our attention mechanism becomes more effective when a query type contains more neighboring nodes in an intersection structure. This indicates that the attention mechanism as well as the original KG training phase are effective in discriminating the correct answer from  \textit{misleading} answers.

\vspace{-0.5cm}
\section{Conclusion} \label{sec:con}
In this work we propose an end-to-end attention-based logical query answering model called contextual graph attention model (\model) which can answer complex conjunctive graph queries based on two geometric operators: the projection operator and the intersection operator. We utilized multi-head attention mechanism in the geometric intersection operator to automatically learn different weights for different query paths. The original knowledge graph structure as well as the sampled query-answer pairs are used jointly for model training. We utilized three datasets (Bio, \db, and \wiki)  to evaluate the performance of the proposed model against the baseline. The results show that our attention-based models (which are trained additionally on KG structure) outperform the baseline models (particularly on the hard negatives) despite using less parameters. The current model is utilized in a transductive setup. In the future, we want to explore ways to use our model in a inductive learning setup. Additionally,  conjunctive graph queries are a subset of SPARQL queries which do not allow disjunction, negation, nor filters. They also require the predicates in all query patterns to be known. In the future, we plan to investigate models that can relax these restrictions.

\bibliographystyle{ACM-Reference-Format}
\bibliography{reference}

\end{document}